\documentclass{article}

\usepackage{PRIMEarxiv}

\usepackage[utf8]{inputenc} % allow utf-8 input
\usepackage[T1]{fontenc}    % use 8-bit T1 fonts
\usepackage{hyperref}       % hyperlinks
\usepackage{url}            % simple URL typesetting
\usepackage{booktabs}       % professional-quality tables
\usepackage{amsfonts}       % blackboard math symbols
\usepackage{nicefrac}       % compact symbols for 1/2, etc.
\usepackage{microtype}      % microtypography
\usepackage{lipsum}
\usepackage{fancyhdr}       % header
\usepackage{graphicx}       % graphics

\usepackage{caption}

\usepackage{subfigure}
\usepackage{tabularx}
\usepackage{multirow}
\usepackage{algorithm}
\usepackage{algorithmic}
\usepackage{amsmath}
\usepackage{xspace}

\usepackage{relsize}
\usepackage{mathrsfs}
\usepackage{makecell}

\usepackage{xcolor,colortbl}
\definecolor{Gray}{gray}{0.85}

\usepackage{natbib}

\definecolor{Gray}{gray}{0.85}
\newcommand{\ours}{\textsc{ShuttleSHAP}\xspace}

\title{ShuttleSHAP: A Turn-Based Feature Attribution Approach for Analyzing Forecasting Models in Badminton}

\author{
  Wei-Yao Wang$^{\dagger}$ \\
  National Yang Ming Chiao Tung University \\
  Hsinchu, Taiwan \\
  \texttt{sf1638.cs05@nctu.edu.tw} \\
\And
  Wen-Chih Peng \\
  National Yang Ming Chiao Tung University \\
  Hsinchu, Taiwan \\
  \texttt{wcpengcs@nycu.edu.tw} \\
\And
  Wei Wang \\
  University of California, Los Angeles \\
  Los Angeles, USA \\
  \texttt{weiwang@cs.ucla.edu} \\
\And
  Philip S. Yu \\
  University of Illinois Chicago \\
  Chicago, USA \\
  \texttt{psyu@uic.edu} \\
}

\begin{document}
\maketitle

% \blfootnote{$^{\dagger}$This work was done during a visiting researcher at UCLA.}
\def\thefootnote{$\dagger$}\footnotetext{This work was done during a visiting researcher at UCLA.}\def\thefootnote{\arabic{footnote}}

\begin{abstract}
Agent forecasting systems have been explored to investigate agent patterns and improve decision-making in various domains, e.g., pedestrian predictions and marketing bidding.
Badminton represents a fascinating example of a multifaceted turn-based sport, requiring both sophisticated tactic developments and alternate-dependent decision-making.
Recent deep learning approaches for player tactic forecasting in badminton show promising performance partially attributed to effective reasoning about rally-player interactions.
However, a critical obstacle lies in the unclear functionality of which features are learned for simulating players' behaviors by black-box models, where existing explainers are not equipped with \textit{turn-based} and \textit{multi-output} attributions.
To bridge this gap, we propose a turn-based feature attribution approach, \ours, for analyzing forecasting models in badminton based on variants of Shapley values.
\ours is a model-agnostic explainer that aims to quantify contribution by not only temporal aspects but also player aspects in terms of multifaceted cues.
Incorporating the proposed analysis tool into the state-of-the-art turn-based forecasting model on the benchmark dataset reveals that it is, in fact, insignificant to reason about past strokes, while conventional sequential models have greater impacts.
Instead, players' styles influence the models for the future simulation of a rally.
On top of that, we investigate and discuss the causal analysis of these findings and demonstrate the practicability with local analysis\footnote{The code will be available in the camera-ready version.}.
\end{abstract}

% % keywords can be removed
% \keywords{Turn-based explainer \and Multi-output attribution \and Sports analytics}

\section{Introduction}
\label{sec:introduction}

% broad impact
The research of applying deep neural networks to simulate patterns has broad applicability across various domains.
Whether it involves investigating previous and future behaviors for sports analytics \cite{DBLP:journals/corr/abs-2211-12217}, financial markets \cite{DBLP:conf/pkdd/LienLW22}, medical simulators \cite{DBLP:journals/cmig/KaramiLR23}, or pedestrian predictions \cite{Xu_2023_CVPR}, these scenarios can be effectively framed as agent forecasting systems characterized by intricate interactions and essential decision-making processes.
Automatically predicting agents' patterns is crucial for improving the effectiveness of decision-making and goal planning.

\begin{figure*}
    \centering
    \includegraphics[width=\linewidth]
    {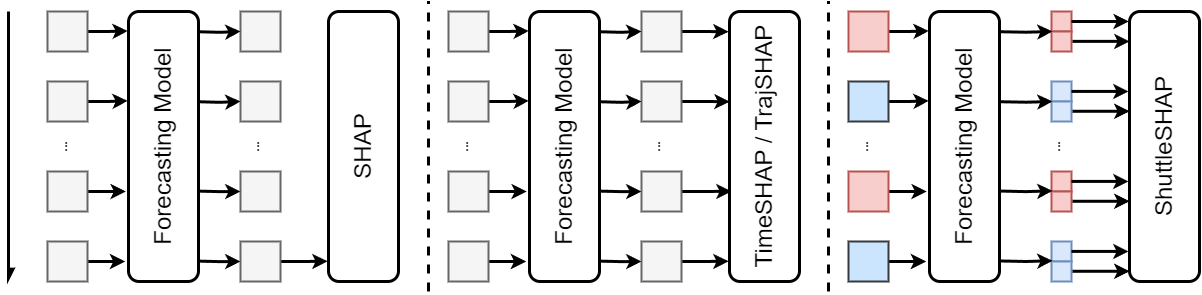}
    \caption{Comparisons of SHAP \cite{DBLP:conf/nips/LundbergL17}, TimeSHAP \cite{DBLP:conf/kdd/0002SCFB21} and TrajSHAP \cite{DBLP:conf/iclr/MakansiKLGJBS22}, and our proposed \ours. Each square denotes a component. Red and blue correspond to different players in a turn-based sequence, where existing works (left and middle parts) view them as the same entities. Two components in a timestamp in the red part represent the shot type and area, where existing explainers fail to incorporate multiple components.}
    \label{fig:differences}
\end{figure*}

% turn-based analytics -> badminton -> forecasting models -> cons
Over the past few years, turn-based sports analytics for action valuing \cite{DBLP:conf/icdm/WangCYWFP21,DBLP:journals/percom/GhoshRCR22,10.1145/3551391}, action detection \cite{DBLP:conf/cvpr/LiuW22,DBLP:journals/sensors/LiLAQC23}, and strategy development \cite{DBLP:conf/cikm/Wang22,Huang_Hsueh_Chien_Wang_Wang_Peng_2023} have attracted researchers due to the sophisticated characteristics that are composed of two players returning strokes alternatively to form a rally, and rapid changes in players' tactics within a match or even a rally.
Therefore, we focus on turn-based sports and adopt badminton as an illustration example since there are no public stroke-level datasets in other turn-based sports \cite{DBLP:conf/aaai/WangSCP22}.
Recently, \cite{DBLP:conf/aaai/WangSCP22} proposed the stroke forecasting task for badminton matches with a promising performance by the turn-based designs of not only rally but also player encoders.
Despite the impressive results, it is notoriously challenging to understand how such turn-based forecasting networks make predictions, and the potential for performance improvements \cite{ShuttleSet22}, and therefore they are often viewed as black-box models.
Understanding the behaviors of black-box models is crucial in high-stakes decisions, especially in sports, given their multifaceted nature \cite{DBLP:conf/kdd/SunDSL20}.
These concerns significantly prevent coaches and practitioners from the development of forecasting models to enhance decision-making in physical training and pre-match tactical investigation.

% challenges of existing explainable methods
To mitigate the aforementioned shortcomings of existing black-box models and potential directions for further progress, we focus on model-agnostic and post-hoc explainers since our goal is to investigate the behaviors from the existing methods for turn-based analytics.
A potential solution is to apply a game theory-based approach, Shapley values (SHAP) \cite{DBLP:conf/nips/LundbergL17}.
% , which has been also adopted in different sports, e.g., basketball \cite{DBLP:journals/anor/MetuliniG23}.
However, SHAP only measures instance-wise feature contributions, which is not applicable to sequential decision-making tasks.
\cite{DBLP:conf/kdd/0002SCFB21} proposed TimeSHAP to explain the predictions on fraud detection tasks, and \cite{DBLP:conf/iclr/MakansiKLGJBS22} extended SHAP to the trajectory prediction task (named TrajSHAP for short).
% by extending SHAP to accommodate both feature- and temporal-wise attributions.
Despite the progress, the temporal-wise explainer is insufficient for turn-based sports analytics due to the compositions of forming a sequence (i.e., rally) alternatively by two players instead of typical sequences with the same targets.
That is, TimeSHAP and TrajSHAP fail to capture players' attributions within a rally as illustrated in Figure \ref{fig:differences}.
Furthermore, there are multiple outputs of each stroke (i.e., shot types and area coordinates); adopting existing explainers suffers from the single-performance quantification of each prediction of the model.
The question, therefore, naturally arises: \textit{Is there a model-agnostic explainer that can further attribute players' contributions in terms of multifaceted performance?}

% our solution
In light of the above challenges, we propose \ours, a turn-based feature attribution framework, to analyze forecasting models in badminton.
Specifically, \ours incorporates Shapley values \cite{shapley1997value,DBLP:conf/nips/LundbergL17} to develop a model-agnostic feature attribution method for turn-based stroke forecasting models.
Moreover, the player influence attribution is proposed by aggregating the contributions of players in an alternative-playing sequence to quantify how well the model leverages player information for forecasting future behaviors.
In this manner, \ours is able to measure the contributions not only from the temporal perspective (i.e., \textit{how do past strokes influence the predictions}) but also from the player perspective (i.e., \textit{how do players influence the predictions}) in terms of multi-faceted performance (i.e., \textit{the performance of shot types and area coordinates}).

% findings
By leveraging our analysis of the representative state-of-the-art turn-based forecasting model ShuttleNet \cite{DBLP:conf/aaai/WangSCP22} and the best-performing sequential model TF \cite{DBLP:conf/icpr/GiuliariHCG20} (observed in \cite{DBLP:conf/aaai/WangSCP22}), we unearthed that 1) Turn-based behaviors do play a critical role in both models in terms of the performance of shot types and area coordinates; 2) Past strokes contribute insignificantly to the area performance and shot type performance for ShuttleNet, while TF without turn-based designs\footnote{In this paper, turn-based design refers not explicitly consider player styles in the model, following \cite{DBLP:conf/aaai/WangSCP22}.} predicts future behaviors based on past strokes.
We, therefore, further delve into the causal analysis of the confounding effects between these approaches and playing strategies in badminton matches.

% contributions
In brief, our contributions are three-fold:
\begin{itemize}
    \item We address, for the first time, feature attribution for the turn-based forecasting problem (\ours) to break new ground for researchers and domain experts to investigate the actual cues about the model performance. \ours is a model-agnostic explainer that can be applied to different forecasting models and can be extended to other turn-based sports (e.g., tennis).
    \item \ours reinforces the capability to quantify the contributions of not only the temporal aspect but also the player aspect. Moreover, it supports multi-output performance attribution for sequential prediction tasks.
    \item We uncovered that the turn-based model for the stroke forecasting task does not significantly utilize past strokes for area predictions, potentially due to the causal intervention design, whereas conventional sequential approaches suffer from confounding effects. In addition, player interactions have important cues for reasoning future strokes for both approaches.
\end{itemize}
\section{Preliminaries}
\label{sec:related-work}

\sloppy
\subsection{The Stroke Forecasting Task}
\label{preliminary-task}
\paragraph{Problem Definition}
A turn-based rally in badminton is denoted as $R = [r_1, r_2, \cdots, r_{|R|}]$ consisting of multiple strokes, where $r_i=(p_i, s_i, a_i)$ is the $i$-th stroke composed of one of two players returning alternatively (A denotes the served player and B otherwise, e.g., $[p_1^A, p_2^B, \cdots, p_{|R|-1}^A, p_{|R|}^B]; p_i \in P$)\footnote{$i$ is omitted to denote players for simplicity in the following.}, the shot type $s_{i} \in S$, and the 2D area coordinates representing the destination locations $a_i \in \mathbb{R}^2$.
The stroke forecasting problem is defined as forecasting the future strokes consisting of shot types and area coordinates $R_{\tau+1:|R|}=[(s_{\tau+1}, a_{\tau+1}), \cdots, (s_{|R|}, a_{|R|})]$ by giving the previous $\tau \in \{2,4,8\}$ strokes $R_{1:\tau}=[(s_{1}, a_{1}), \cdots, (s_{\tau}, a_{\tau})]$ and the two players for each rally.
Generally, simulating players' future behaviors requires models to learn past situations in a rally and the styles of both players.

\begin{figure*}
    \centering
    \includegraphics[width=0.83\linewidth]
    {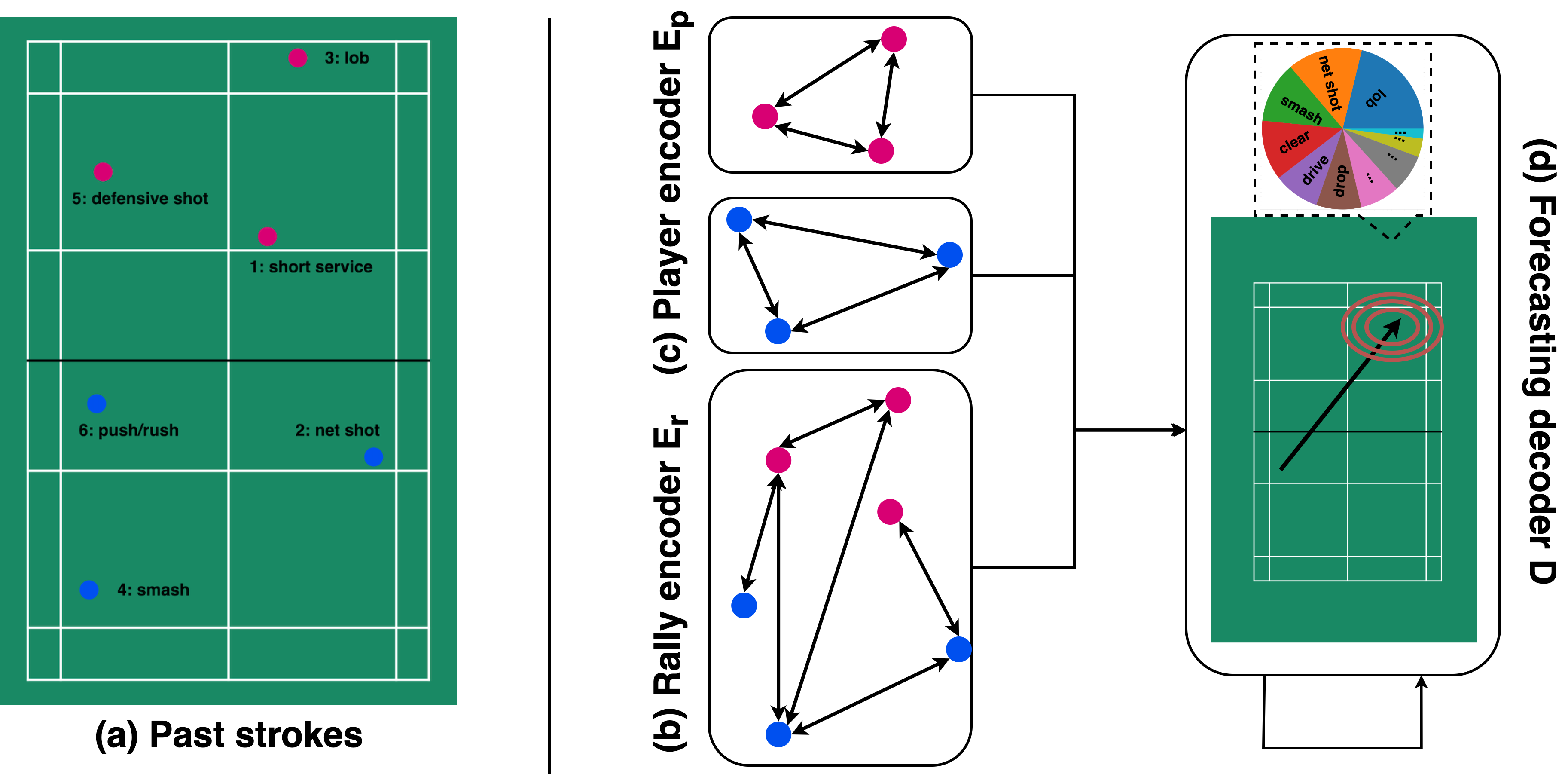}
    \caption{General framework of existing turn-based forecasting approaches: Given the past $\tau$ strokes (a), the rally encoder (b) captures the rally situation as the rally embeddings, and the player encoder (c) captures the overall styles of both players. Afterwards, the forecasting decoder (d) predicts future strokes autoregressively. Note that conventional sequence prediction models (e.g., TF \cite{DBLP:conf/icpr/GiuliariHCG20}) do not have the player encoder to explicitly consider players' styles.}
    \label{fig:architecture}
\end{figure*}

\paragraph{Existing Turn-Based Forecasting Models}
Figure 2 (b)-(d) summarizes the general framework that unifies the state-of-the-art turn-based forecasting models and most conventional sequence forecasting methods.
Given past strokes of the rally (a):
\begin{itemize}
    \item[(b)] A rally encoder $E_r$ that captures previous strokes in a rally $H_r = E_r(R_{1:\tau})$.
    \item[(c)] A player encoder $E_p$ that learns the overall styles of both players separately as $H_p^A, H_p^B = E_p(R_{1:\tau}^A, R_{2:\tau}^B)$, where $R_{1:\tau}^A=[(p^A, s_1, a_1), (p^A, s_3, a_3), \cdots]$ and $R_{2:\tau}^B=[(p^B, s_2, a_2), (p^B, s_4, a_4), \cdots]$.
    \item[(d)] A forecasting decoder $D$ that predicts future strokes based on the outputs of (b) and (c), and continues the process autoregressively $R_{\tau+1:|R|} = D(H_r, H_p^A, H_p^B)$. 
\end{itemize}
It is noted that each shot type, the area coordinates, and player information are first transformed to $d$-dimensional embeddings, respectively.
The existing turn-based forecasting approach \cite{DBLP:conf/aaai/WangSCP22} adopts Transformer-based \cite{DBLP:conf/nips/VaswaniSPUJGKP17} extractors for $E_r$ and $E_p$, respectively, and then fuses them with a position-aware technique.
As the stroke forecasting task remains unexplored, a best-performing sequential prediction model \cite{DBLP:conf/icpr/GiuliariHCG20} is also employed following the original paper to investigate the behaviors compared with turn-based models, where the model does not contain a player encoder and the fusion technique; that is, $R_{\tau+1:|R|}=D(H_r)$.

\subsection{Model-Agnostic Explainers}
Model-agnostic explainable methods have pioneered interpretability in deep learning models due to their flexibility to plug into already trained models.
A feature is viewed as important if shuffling its value changes the performance, since it implies that the model relies on the feature for predictions, while it is considered unimportant if unchanged due to the negligence of the feature for making the prediction.
Earlier methods are mainly divided into two groups: perturbation-based \cite{DBLP:journals/corr/LiMJ16a,feng-etal-2018-pathologies} and gradient-based \cite{DBLP:journals/corr/SmilkovTKVW17,DBLP:conf/acl/HanWT20} explainers.
The former perturbs input features for computing relevant differences compared with the original ones, and the latter introduces noises into the input.
However, perturbation-based explainers violate the \textit{sensitivity axiom}, while gradient-based explainers violate the \textit{implementation invariance axiom}, i.e., the attributions are always identical for two functionally equivalent networks \cite{DBLP:conf/icml/SundararajanTY17}.

\cite{DBLP:conf/nips/LundbergL17} introduced SHAP\footnote{More details about SHAP are discussed in Appendix \ref{appendix-discussion}.}, which is the only method that fulfills these desirable axioms based on Shapley values \cite{shapley1997value}, that maps an input $N = \{r_1, \cdots, r_n\}$ to a game, where players are the individual features $r_i$ and the payout is the model behavior (e.g., prediction or performance).

Formally, the contributions of the individual feature $r_i$ can be quantified by its Shapley values $\phi(r_i)$:
\begin{equation}
    \phi(r_i) = \sum_{S \subseteq N \setminus \{r_i\}} \frac{|S|!(|N|-1-|S|)!}{|N|!} (f(S \cup \{r_i\}) - f(S)),
\end{equation}
where $S$ is a subset of features that exclude the $i$-th feature and $f$ represents the model behavior.
Our proposed \ours is built on top of SHAP not only to enable both temporal and player aspects but also to preserve these axioms.
\section{The Proposed \ours}
\label{sec:method}

\subsection{Prerequisite}
\paragraph{What model behavior $f$ is to be attributed?}
\label{model-behavior}
We are interested in the improvement of understanding which shot types and area coordinates are used by models to perform well; that is, quantifying the contribution of each input stroke to the \textit{performance} of a given model, which is also a common choice of observing model behaviors, e.g., \cite{DBLP:conf/iclr/MakansiKLGJBS22}.
Therefore, the behavior of $f$ is defined as the relevant error of the output prediction:
\begin{equation}
    f := \mathcal{L}(\hat{R}_{\tau+1:|R|}, R_{\tau+1:|R|}),
\end{equation}
where $\mathcal{L}$ can be auxiliary losses.

Following the setting of the original task, $\mathcal{L}$ includes cross-entropy loss (CE) $\mathcal{L}_{type}$ for shot type prediction and negative log-likelihood loss (NLL) $\mathcal{L}_{area}$ for area coordinate prediction:
\begin{equation}
  \mathcal{L}=\mathcal{L}_{type}+\mathcal{L}_{area},
  \label{total_loss}
\end{equation}
\begin{equation}
    \mathcal{L}_{type} = CE(s_i, \hat{s}_i),
    \label{shot_loss}
\end{equation}
\begin{equation}
  \mathcal{L}_{area}=NLL(x_i, y_i, \hat{x}_i, \hat{y}_i).
  \label{area_loss}
\end{equation}
In this manner, it is able to quantify the microscopic attributions with respect to multifaceted predictions and the macroscopic attribution of a stroke.
To provide the macroscopic attribution, the scores from shot type prediction and area coordinate prediction are averaged to view both outputs as having the same importance.

% mention how to exclude features -> reference baseline
\paragraph{How to design dropped features for $S$?}
Since the exact computation of Shapley values requires exponential computation, a common solution for efficiency is to replace the dropped features with the reference value, e.g., set as zero \cite{DBLP:conf/icml/AnconaOG19}.
However, setting the reference value for the dropped features to either random/mean values from the dataset or zero padding values introduces some unexpected signals to represent the process of a badminton rally (e.g., always returning the stroke to the top-left corner with a padded shot type).

To design a reasonable reference to represent the dropped features for stroke forecasting, the reference stroke is introduced to be a \textit{defensive stroke returning to the middle of the half-court}, which is a flexible option that can defensively return most strokes back to the opponent for simulating their tactics:
\begin{equation}
    (s_i, a_i) \leftarrow (s', a'),
\end{equation}
where the $i$-th shot type is replaced with a defensive shot $s'$, and the $i$-th area coordinates are replaced with the coordinates in the middle of a half-court $a'$.
This modified version of Shapley values enables us to quantify the strokes that we aim to investigate relative to the consistent single-reaction strategy, and to apply this technique to different existing models.
The proposed design of reference values is generic to be adapted to other turn-based sports (e.g., tennis); that is, setting the reference values as the defensive shot that is more likely to be returned.

It is noted to acknowledge that the justification for quantifying its contribution through the extent of predictive enhancement is not always valid, contrary to prevailing perceptions. \cite{janzing2013quantifying} (Section 3.3 and Example 7) argued that evaluating causal potency by omitting the feature for prediction is fundamentally flawed and demonstrated that it should be randomized instead.
Therefore, the ablative experiments of randomizing dropped features are excluded in this paper since it is inconclusive based on flawed modifications.

\subsection{Past Strokes Attributions}
% motivation
The real-time decision-making of players based on the immediate situation is considered by their instant reactions for returning to the next stroke instead of the strokes far from now \cite{macquet2007naturalistic}, which piques our curiosity about how the models take advantage of past information for future simulations.
Note that \textit{past} strokes refer to the first $\tau$ strokes.
With the foundation of \ours, quantifying the contribution of past strokes in terms of multi-performance is possible, while existing explainers fail to attribute multiple outputs in a timestamp.

% how to compute for a stroke
To address this problem, the shot type is replaced with the defensive shot, and the area coordinates are changed to the middle of a half-court from the second stroke:
\begin{equation}
\label{past-replace}
    R_{2:\tau} \leftarrow [(p_2, s', a'), \cdots, (p_\tau, s', a')].
\end{equation}
The reason to retain the first stroke is that the first stroke is the service shot, which has limited choices: a short service to the half-middle court or a long service to the backcourt, while it is impossible to be a defensive shot to the middle court due to the serving area rule\footnote{The illustration rule is presented in Appendix A.}.
Therefore, the first stroke is contained in a rally to not only reflect the proper starting stroke but also to remove noise from the simulation.

% how to aggregate to form a score
After obtaining the contribution of each stroke, aggregating these Shapley values over several strokes is essential to draw conclusions about the model behavior on the turn-based dataset.
Formally, given the Shapley values of a stroke in a rally $\phi(r_i)$, the aggregated value of a rally $\phi_R^P$ is performed as:
\begin{equation}
    \phi_R^P = \rho(\phi_{r_i \in R_{\tau+1:|R|}}(r_i)),
\end{equation}
where $\rho$ denotes the aggregation function.

The potential selection of $\rho$ is to adopt the max operation to choose the most important timestamp for each sequence, similar to \cite{DBLP:conf/iclr/MakansiKLGJBS22}; this is problematic when measuring multiple strokes with true causal influences throughout different scenarios.
For instance, some transition strokes are returned to organize their tactics, which are still useful for providing explanations for simulation.
Therefore, we opt for the average operation for $\rho$ to represent the Shapley values of a rally in terms of the past strokes attributions.
For the global aggregation (i.e., the entire dataset), the average operator is also incorporated.
A higher non-negative value suggests that the previous strokes positively contribute to predicting future strokes in general, while a zero score indicates that none of the past strokes are used for predictions.

% \begin{figure*}[t]
%     \centering
%     \subfigure[Shapley values based on shot type performance (CE).]{
%         \includegraphics[width=.43\linewidth]{figures/shap_all_ce.png}
%     }
%     \subfigure[Shapley values based on area performance (NLL).]{
%         \includegraphics[width=.43\linewidth]{figures/shap_all_nll.png}
%     }
%     \caption{Shapley values in three different given strokes of ShuttleNet and TF on the stroke forecasting benchmark, where the numbers of given strokes are denoted in the subscript of each model.}
%     \label{fig:shap_all}
% \end{figure*}

\subsection{Player Influence Attributions}
% motivation
In addition to temporal attributions, one of the captivating characteristics of badminton is multiple alter-returning agents in a sequence, which is different from the conventional sequential tasks composed of a single agent in an entire sequence.
This is considered in the state-of-the-art model, ShuttleNet \cite{DBLP:conf/aaai/WangSCP22}, which motivates us to propose the player influence attribution technique aiming to examine to what extent both players contribute to the predictive performance.

% how to compute for a stroke
For each target player, the same dropping strategy of the past stroke attributions is adopted to replace the shot type and the area coordinates with the defensive shot and the middle of the half-court, respectively.
Furthermore, the target player of each stroke is changed to the other player, e.g., $p^A$ is modified as $p^B$ if the target player is A and vice versa.
Formally, the replacements are defined as follows:
\begin{equation}
\label{A-replace}
    R_{3:|R|}^A \leftarrow [(p^B, s_3', a_3'), (p^B, s_5', a_5'), \cdots],
\end{equation}
\begin{equation}
\label{B-replace}
    R_{2:|R|}^B \leftarrow [(p^A, s_2', a_2'), (p^A, s_4', a_4'), \cdots].
\end{equation}
This imputation makes the ever-changing player act as a static serving machine against himself/herself, where the relevance error compared with the performance of the original one can then be used to represent the influence attribution of the players.
Note that Eq. (\ref{A-replace}) and Eq. (\ref{B-replace}) are executed independently to obtain the attributions of both players, and the first stroke retains the same due to the serving stroke.

% how to aggregate to form a score
To perform contribution aggregation, the attributions of multiple Shapley values are first averaged for each player of each stroke, and the attributions of two players of a stroke are then aggregated since we hypothesize that players consider that their own styles and those of their opponents have equal importance when returning the next stroke.
Afterwards, the same procedure used for past stroke attributions is adopted to obtain the player influence attributions, considering the same importance over multiple strokes.
\begin{figure*}[t]
    \centering
    \includegraphics[width=\linewidth]{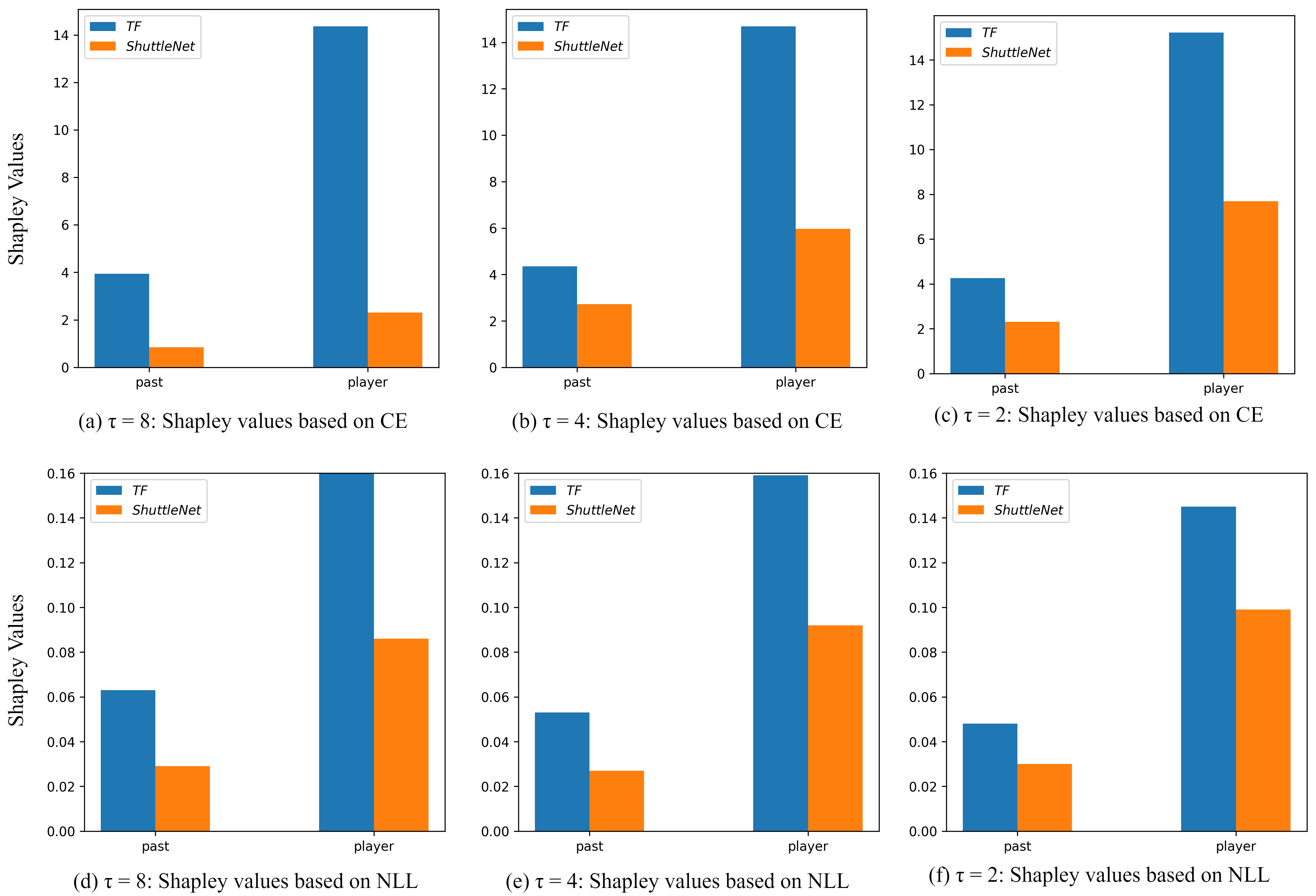}
    \caption{Shapley values in three different given strokes ($\tau$) of ShuttleNet and TF on the stroke forecasting benchmark, where the x-axis denotes the attribution targets.}
    \label{fig:shap_all}
    \vspace{-7pt}
\end{figure*}

\section{Experiments and Analysis}
\label{sec:experiments}

\subsection{Experimental Setup}
\paragraph{Benchmarks.}
As there is no existing forecasting work tackling other turn-based sports as discussed in Section \ref{sec:introduction}, \ours is employed on a state-of-the-art turn-based model, ShuttleNet \cite{DBLP:conf/aaai/WangSCP22}, and a best-performing conventional sequential model, TF \cite{DBLP:conf/icpr/GiuliariHCG20}, as introduced in Section \ref{preliminary-task} to incorporate the badminton dataset with stroke-level records \cite{DBLP:conf/aaai/WangSCP22}.
The dataset contains 31 professional men's singles and women's singles players ($P$) playing 75 matches, which includes 180 sets, 4,325 rallies, and 43,191 strokes.
The set of shot types $S$ consists of \textit{
  clear, % 長球
  net shot, % 放小球
  smash, % 殺球
  push/rush, %推撲球
  drop, % 切球
  drive, % 平球
  lob, % 挑球
  defensive shot, %接殺防守
  short service, % 發短球
}and \textit{long service}, which are defined by \cite{10.1145/3551391}.
We follow the same splitting method to split the dataset for training and testing.
The hyper-parameters of both models are implemented following the original paper detailed in Appendix B.

\paragraph{Evaluation Metrics.}
Following \cite{DBLP:conf/aaai/WangSCP22}, we adopt cross-entropy (CE) for the shot type measurement, and mean square error (MSE) and mean absolute error (MAE) are used for the area coordinates measurements.
As discussed in Section \ref{model-behavior}, CE and NLL are used for reporting the contributions of models.
Three different given strokes $\tau=8, 4$, and $2$ following the stroke forecasting task are conducted to investigate different behaviors in terms of model predictions.

\begin{table*}
    \centering
    \caption{Retrained performance of TF and ShuttleNet on modified player information.}
    \begin{tabular}{l|ccc|ccc|cccccc}
    \toprule
    & \multicolumn{3}{c|}{$\tau=8$} & \multicolumn{3}{c|}{$\tau=4$} & \multicolumn{3}{c}{$\tau=2$} \\
    \cmidrule{2-10}
    Model & CE & MSE & MAE & CE & MSE & MAE & CE & MSE & MAE \\
    \midrule
    TF         & 2.4398 &  1.7125 &  1.4247 & 2.4392 &  1.6792 &  1.3974 & 2.4925 &  1.7102 &  1.4116 \\
    w/o player & 1.9951 &  2.8248 &  1.8980 & 1.9833 &  2.7121 &  1.8643 & 1.9869 &  2.8332 &  1.9008 \\
    \rowcolor{Gray}
    Difference & 0.4447 & -1.1123 & -0.4733 & 0.4559 & -1.0329 & -0.4669 & 0.5056 & -1.1230 & -0.4892 \\
    \midrule
    ShuttleNet &  1.9854 &  1.5950 &  1.3993 & 2.0568 &  1.6101 &  1.3954 & 2.1107 &  1.6140 &  1.3960 \\
    w/o player &  2.0001 &  2.7777 &  1.8840 & 1.9818 &  2.7549 &  1.8733 & 1.9856 &  2.4514 &  1.7931 \\
    \rowcolor{Gray}
    Difference & -0.0147 & -1.1123 & -0.4733 & 0.0750 & -1.1448 & -0.4779 & 0.1251 & -0.8374 & -0.3971 \\
    \bottomrule
\end{tabular}
    \label{tab:diff-Q2}
\end{table*}

\subsection{Quantitative Results}
Figure \ref{fig:shap_all} summarizes the contributions in terms of shot type performance (CE) and area performance (NLL) for the player influences (player) and for the past strokes (past).
It is worth noting that the performance of all models (i.e., the first row of each model in Tables \ref{tab:diff-Q2} and \ref{tab:diff-Q1}) is similar to the results in the original paper.

\paragraph{A1: Analysis on Player Influences.}
From the results of \textit{player} influences, it is observed that the attributed values of players have significant impacts in terms of the shot type (CE) and area predictions (NLL) for TF and ShuttleNet compared with the \textit{past} stroke attributions.
In addition, ShuttleNet highly depends on the player styles when the number of given past strokes reduces (orange), while TF decreases the contributions for predicting future landing locations (blue).
These behaviors suggest that turn-based behaviors indeed play a critical role in simulating future strokes including shot types and landing locations.

\begin{table*}
    \centering
    \caption{Retrained performance of TF and ShuttleNet on different past strokes.}
    \begin{tabular}{l|ccc|ccc|cccccc}
    \toprule
    & \multicolumn{3}{c|}{$\tau=8$} & \multicolumn{3}{c|}{$\tau=4$} & \multicolumn{3}{c}{$\tau=2$} \\
    \cmidrule{2-10}
    Model & CE & MSE & MAE & CE & MSE & MAE & CE & MSE & MAE \\
    \midrule
    TF         &  2.4398 & 1.7125 &  1.4247 &  2.4392 & 1.6792 & 1.3974 & 2.4925 & 1.7102 & 1.4116 \\
    w/o past   &  2.4813 & 1.6870 &  1.4143 &  2.4496 & 1.6358 & 1.3846 & 2.4274 & 1.7015 & 1.4132 \\
    \rowcolor{Gray}
    Difference & -0.0415 & 0.0255 &  0.0104 & -0.0104 & 0.0434 & 0.0128 & 0.0651 & 0.0087 & -0.0016 \\
    \midrule
    ShuttleNet &  1.9854 & 1.5950 & 1.3993 &  2.0568 & 1.6101 & 1.3954 &  2.1107 &  1.6140 & 1.3960 \\
    w/o past   &  1.9924 & 1.5949 & 1.3953 &  2.0697 & 1.5830 & 1.3939 &  2.1249 &  1.6216 & 1.3939 \\
    \rowcolor{Gray}
    Difference & -0.0007 & 0.0001 & 0.0040 & -0.0129 & 0.0271 & 0.0015 & -0.0142 & -0.0076 & 0.0021 \\
    \bottomrule
\end{tabular}
    \label{tab:diff-Q1}
\end{table*}

To further evaluate our findings, we retrain these models from scratch without considering the styles of players.
This can be achieved by imputing one player and the corresponding strokes with Eq. (\ref{A-replace}) or Eq. (\ref{B-replace}) in the training set respectively, and then average the performance results.
Table \ref{tab:diff-Q2} reports the original and the retrained results of two models with different past strokes (denoted as w/o player).
Similar to the analysis, the results indicate that both models show substantial differences with and without player influences, suggesting that players' styles influence the predictions.
Notably, the performance of shot type predictions for both models can be improved when neglecting the considerations of players' styles, which leaves a potential direction to tackle the multi-output turn-based sequences.

\paragraph{A2: Analysis on Past Stokes.}
The comparison of \textit{player} and \textit{past} stroke attribution in Figure \ref{fig:shap_all} reveals that for both models, shot types and landing locations make a relatively minor contribution compared with player influences.
% both models contribute relatively minor than player influences for shot types and landing locations.
Moreover, the attributions of past strokes expose that ShuttleNet insignificantly takes advantage of past strokes for predicting future strokes compared with TF (a-c).
This implies that sequential models without turn-based designs rely on not only implicit player influences but also previous situations in the rally, while ShuttleNet is able to simulate future behaviors by primarily focusing on player styles.

This result draws our attention to whether the performance of ShuttleNet remains similar if the context of the past strokes is dropped as the contribution of each stroke is quantified to the performance of a given model (Section \ref{model-behavior}).
Therefore, the same process is conducted to observe the relative difference between considering and removing (w/o past) past strokes presented in Table \ref{tab:diff-Q1}, where w/o past is implemented by injecting Eq. (\ref{past-replace}) into the training set.
When discarding the information from past strokes, the performance of ShuttleNet remains stable in terms of all metrics and different given lengths, while the difference of TF impacts more substantially, although not as much, when neglecting player information.
This demonstrates that past strokes are not considered significant in either model, whereas player influences play a vital role in simulating future strokes from encoding past strokes either implicitly or explicitly.
Moreover, it also highlights that ShuttleSHAP is able to identify the contributions of each feature in terms of multifaceted performance and the performance yields consistent with our findings.

\begin{figure}[t]
    \centering
    \subfigure[]{
        \includegraphics[width=.25\linewidth]{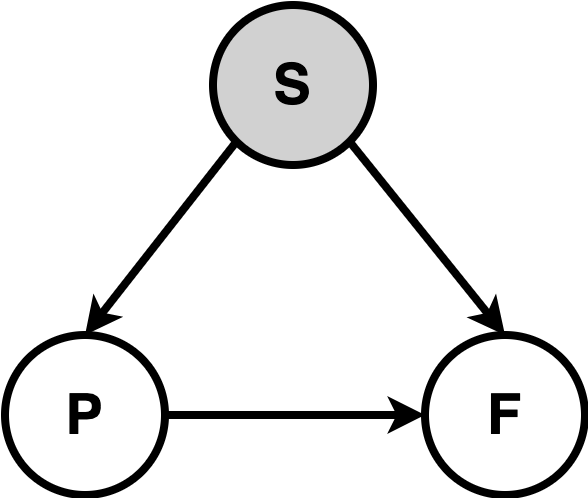}
        \label{fig:confounding-original}
    }
    % \hfill
    \hspace{50pt}
    \subfigure[]{
        \includegraphics[width=.25\linewidth]{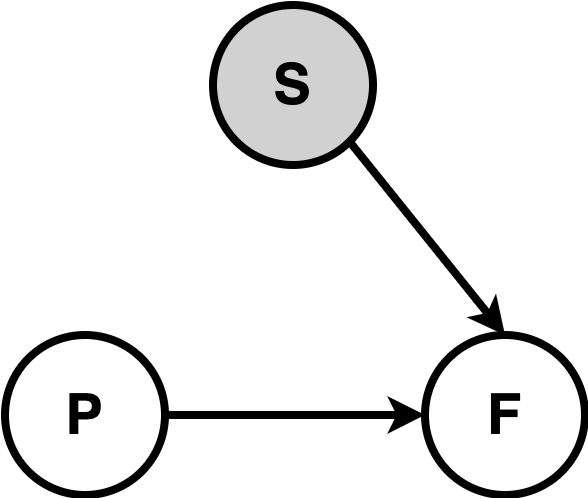}
        \label{fig:confounding-removed}
    }
    \caption{The analysis of the structural causal model for stroke forecasting. (a): The original causal relations, which is the design of the conventional sequential model TF. (b): After the causal intervention, which is the design of the turn-based model ShuttleNet.}
    \label{fig:confounding}
\end{figure}

\subsection{Discussions of the Findings}
From the above results, we aim to delve into the interesting question to provide possible research directions: \textit{Why may ShuttleNet simulate better than conventional sequential models even though it does not much consider past strokes?}
To answer this question, we analyze the true causalities from the perspective of causal intervention, where the spurious correlations between cause and effect are cut off to systematically investigate the problem \cite{pearl2009causality,pearl2016causal}.
As shown in Figure \ref{fig:confounding-original}, the causalities can be formulated among three variables: past strokes (P), future strokes (F), and player styles (S) into a structural causal model \cite{pearl2009causality}.
The directed edges represent causalities between two variables: cause $\rightarrow$ effect, and each edge is depicted as follows:
\begin{itemize}
    \item \textbf{P $\rightarrow$ F: } The future strokes can be inferred by the past strokes. 
    \item \textbf{S $\rightarrow$ F: } The player styles influence the future strokes. For instance, players aim to return the stroke to create an opportunity to take the initiative to seize the rally.
    \item \textbf{S $\rightarrow$ P: } The player styles also affect the past strokes for the same reason with \textbf{S $\rightarrow$ F}.
\end{itemize}

Based on the definition of each edge, we can identify that the role of player styles is the confounder in the stroke forecasting task since there exists a backdoor path \textbf{P $\leftarrow$ S $\rightarrow$ F}, which indicates that the player styles still correlate the past strokes with the future strokes even if some past strokes have little likelihood to simulate some irrational strokes, causing the spurious correlations in the model.
For example, players return strokes passively, possibly because of the long-lasting shots in a rally to preserve energy instead of the aggressive strokes from the opponents.

From Figure \ref{fig:architecture}, the framework design of ShuttleNet eliminates the confounding effect by the causal intervention design \cite{pearl2009causality}, which cuts off the directed edge \textbf{S $\rightarrow$ P} to learn interventional probability and to reduce the prediction errors \cite{DBLP:conf/cvpr/NiuTZL0W21}.
However, conventional sequential models for trajectory prediction without causal intervention (e.g., TF) make the wrong use of player styles implicitly in the stroke forecasting task, leading to inaccurate prediction as raised in \cite{DBLP:conf/aaai/GeSH23}.
Therefore, the confounding effect may be one of the reasons that ShuttleNet achieves promising performance without much considering past strokes since players highly depend on the immediate situation for decision-making instead of the strokes that are from long ago \cite{macquet2007naturalistic}.

\begin{figure*}
    \centering
    \includegraphics[width=\linewidth]{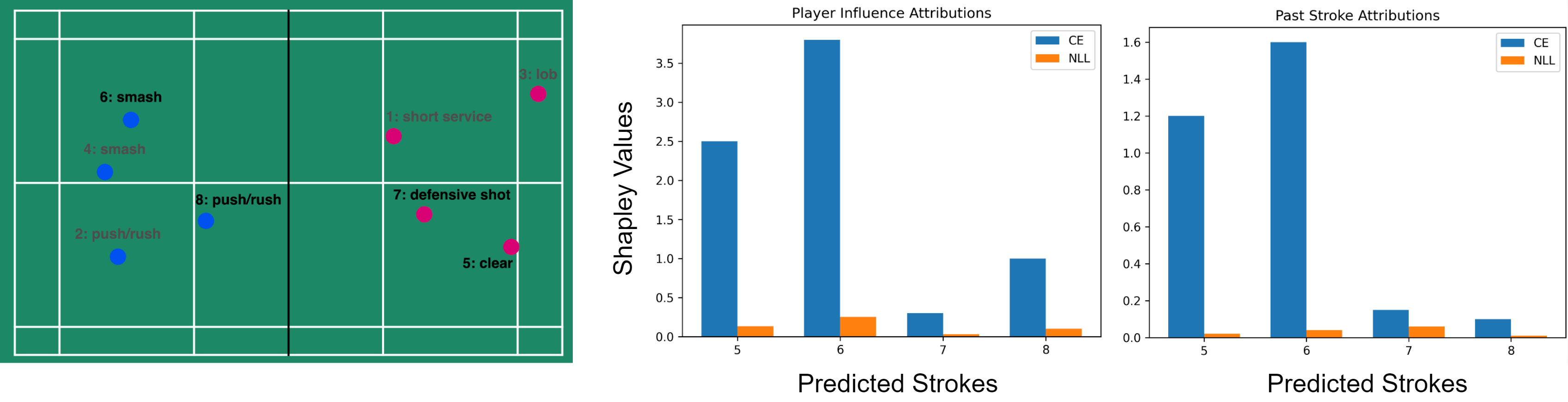}
    \caption{Qualitative analysis of applying \ours on ShuttleNet for a scenario where the styles of players are important for returning the next strokes. The first four strokes in grey are the given strokes. CE/NLL stands for the Shapley values based on the corresponding losses (Eq. \ref{shot_loss} and \ref{area_loss}). In the left figure, the landing locations in the left court are denoted in blue dots and those in the right court are denoted in red dots. In the middle and right figures, the Shapley values from five to eight predicted strokes are illustrated in terms of both shot types (CE) and landing locations (NLL).}
    \label{fig:case}
    \vspace{-8pt}
\end{figure*}

\subsection{Usage of \ours: Local Analysis}
The analysis of the contributions of each stroke in terms of shot type and area predictions can be viewed as understanding the behaviors of the model to help domain experts reduce the concerns when deploying the simulation model for pre-match training and to provide researchers with insights into the model design.
To that end, a paradigm is to investigate the practicability by applying \ours to ShuttleNet with the player influence attributions and past stroke attributions.
Figure \ref{fig:case} demonstrates a real-world scenario from a high-ranking men's singles match where the first four strokes in grey text are given for the model to forecast the future strokes in black text in the left figure.

The model predicts the fifth stroke with a clear to the backcourt after receiving the aggressive smash, and then the player on the right court launches another attack by returning another smash.
Afterwards, the player on the left court has limited option to return and is forced to return the defensive shot to the middle court, and then loses the rally from failing to return back to the push/rush stroke.
ShuttleNet equipped with our method attributes the fifth and sixth strokes significantly to player influences as well as past strokes in terms of shot types and landing locations, which points out that both players return strokes mainly based on their styles.
This can be seen in the third and fourth strokes, where the former is a passive stroke with a lob and the latter is an aggressive stroke with a smash.
Therefore, we can conclude that the left player plays more passively and the right player plays more aggressively.

In addition, the locations of the fourth and sixth strokes are close to the middle court, which also shows that the style of the right player is to attack the middle location.
This behavior is also identified by \ours with the highest attribution in the middle figure.
As the seventh stroke is forced to return, \ours attributes this stroke as insignificant to the player influences for the shot type, but considers it significant for the eighth stroke also due to the aggressive stroke.
With the help of \ours providing quantitative attributions, badminton experts and researchers are able to analyze and interpret forecasting results from the model in terms of microscopic cues (i.e., shot types and landing locations) for the specific rally, and potentially improve pre-match training and tactic investigations.
\vspace{-6pt}
\section{Conclusions}
\vspace{-3pt}
\label{sec:conclusion}
This paper presents \ours, the first feature attribution method for the turn-based forecasting problem.
Distinct from existing model-agnostic explainers, \ours analyzes the actual cues that are used by models to improve forecasting performance not only from the temporal aspect but also from the player aspect.
Meanwhile, it enables multi-output contributions that can quantify shot type as well as area attributions.
Our proposed method quantifies causal influence in the context of the stroke forecasting task for benchmarking how well existing models make predictions based on such influence.
Furthermore, we reveal that the turn-based model does not rely on past strokes for future simulation, most likely due to the causal intervention design, while conventional sequential models learn spurious relations between the player styles, and historical and future trajectories due to the confounding effect.
An analysis scenario of local analysis is illustrated to show the interpretation of real-world simulations.
Since this paper points out insights into existing turn-based models, future work could investigate a more advanced method to improve the turn-based sequence forecasting problem.

\appendix

\section{Definitions in Badminton}

\subsection{Restrictions of Serving Area}
\begin{figure}
    \centering
    \includegraphics[width=0.85\linewidth]{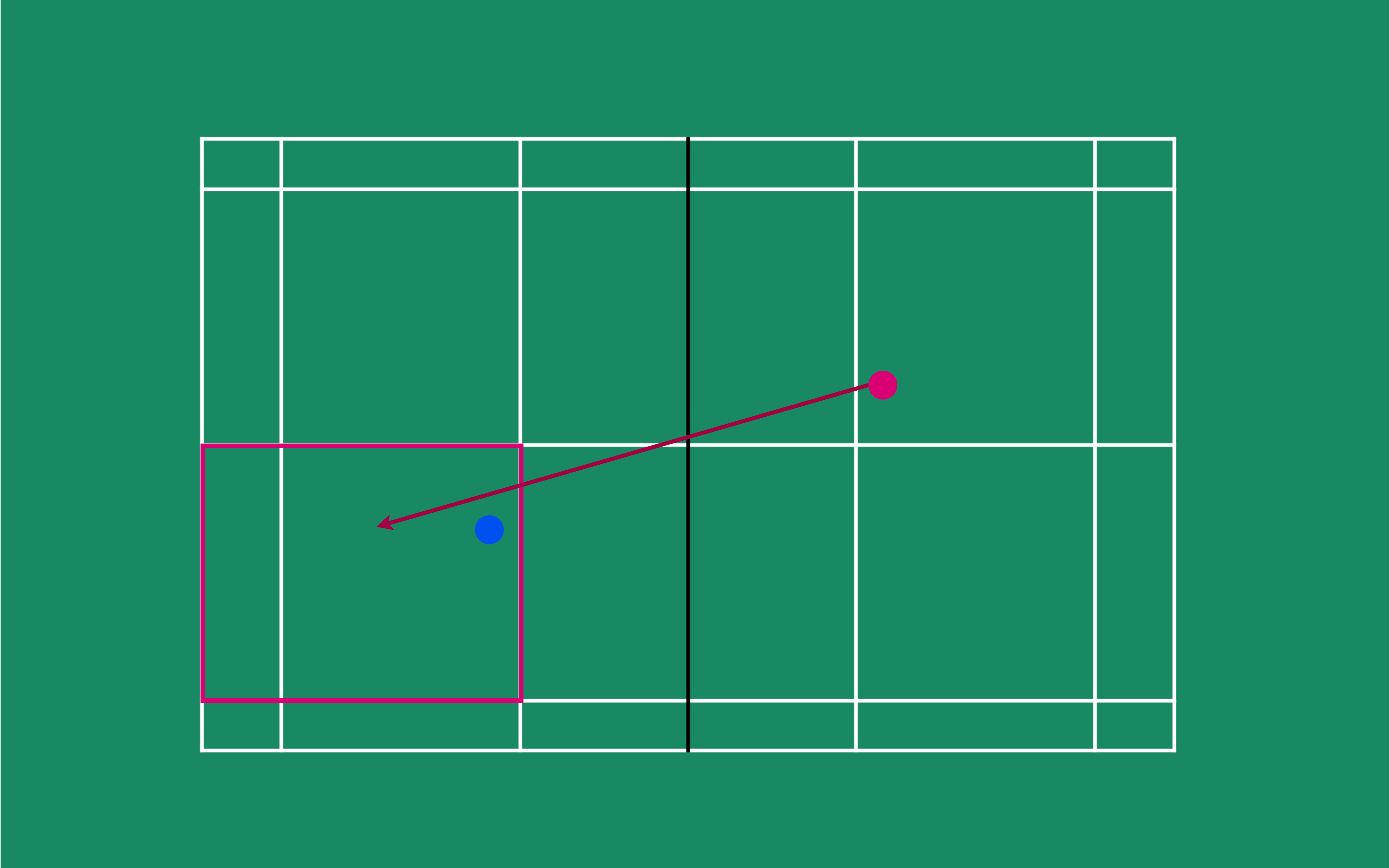}
    \caption{An illustration of the ruled area for the serving stroke.}
    \label{fig:serving-illustration}
\end{figure}

According to the rule of badminton, the serving player needs to serve the shuttle to the left half or right half (depending on the score) of the court as visualized in Figure \ref{fig:serving-illustration}, which limits the option to be manipulated for dropping features.

\subsection{Dataset Collection}
\label{dataset-collection}
Our data are acquired from \cite{DBLP:conf/aaai/WangSCP22}, which were labeled by domain experts with an annotation tool.
The tool provides the interface and standardized protocol with the stroke detection by \cite{9302757} from the scene, which reduces the burden of domain experts for moving each frame to find events of returning strokes. 
They are asked to annotate the shot type and the location of the players by clicking the provided options and the scene of the court.
After annotating a match, several detection cases are launched to automatically examine the quality, e.g., missing detection and irregular score detection. Besides, some rallies are randomly sampled in each match by humans to ensure quality. 
% The annotated matches are reported in Table \ref{tab:match-info}, which contains 51 men's singles and 24 women's singles. 
% The distribution of the rally length is shown in Figure \ref{fig:rally-distribution}, where the max length of the rallies is 62 and the distribution of the shot type is demonstrated in Figure \ref{fig:shot-type-percent}.
% The descriptions of shot types are summarized in Table \ref{tab:shot-type-descriptions}.
% Parts of the dataset are provided in the supplementary.

% \begin{figure}
%   \centering
%   \includegraphics[width=0.92\linewidth]{figures/experiment_rally_distribution.png}
%   \caption{Distribution of rally lengths in the dataset.}
%   \label{fig:rally-distribution}
% \end{figure}

% \begin{figure}
%   \centering
%   \includegraphics[width=0.92\linewidth]{figures/shot-type-percent.png}
%   \caption{Distribution of the shot type in the badminton dataset.}
%   \label{fig:shot-type-percent}
% \end{figure}

% \begin{table}
%   \centering
%   \input{tables/match-info}
%   \caption{Annotated matches in the dataset.}
%   \label{tab:match-info}
% \end{table}

% \begin{table*}
%   \centering
%   \input{tables/shot-type-descriptions}
%   \caption{The description of each shot type in the dataset.}
%   \label{tab:shot-type-descriptions}
% \end{table*}

\section{Implementation Details}
For both models, the dimension of embeddings and contexts ($d$) was set to 32, the number of heads used in multi-head attention and multi-head type-area attention was set to 2, and the inner dimension of feed-forward layer was 64.
The max sequence length of a rally was 35.
The layer normalization and dropout technique with a dropout rate of 0.1 were used for each sublayer.
The batch size was 32 and the number of training epochs was 150 using Adam as the optimizer. The learning rate was set to 0.0001.
To evaluate stochastic models, we generated $K=10$ samples and took the closest one to ground truth for evaluation following the original work.
The input of area coordinates was normalized with the mean as zero. All the training and evaluation phases were conducted on a machine with a Nvidia GTX 3060 GPU.
\section{Further Discussions of Shapley Values}
\label{appendix-discussion}
Generally, there are various variants of Shapley values that mainly focus on the following two design choices:
\begin{itemize}
    \item[1)] What model behavior $f$ is to be attributed?
    \item[2)] How to design dropped features for $S$?
\end{itemize}

In the context of attributing model performance, a frequently chosen initial consideration pertains to the performance of the model itself, a focus also adopted in this paper.
Regarding the second option, SHAP \cite{DBLP:conf/nips/LundbergL17} proposes a methodology wherein dropped features are substituted with samples drawn from the conditional data distribution based on the non-dropped features.
This approach aims to approximate the marginal distribution, thereby simplifying underlying assumptions \cite{DBLP:conf/aistats/JanzingMB20}.
Conversely, \cite{DBLP:conf/aistats/JanzingMB20} contends, from a causal standpoint, that the correct distribution to sample from is the marginal distribution.
This assertion is grounded in the argument that the process of dropping features corresponds naturally to an interventional distribution, rather than a conditional distribution.
Meanwhile, the replacement of dropped features involves utilizing a predefined baseline, as elucidated in \cite{DBLP:conf/icml/SundararajanN20}.

It is crucial to underscore that the relevance of features in forecasting future strokes should not be conflated with their causal impacts; nevertheless, these two facets are intrinsically interconnected \cite{granger1969investigating}. This linkage is substantiated by adhering to the causal Markov condition and causal faithfulness, outlined in \cite{peters2017elements}, wherein three key assumptions are met: 
1) Each feature exerts influence on others only with a nonzero time lag.
2) There are no latent (unobserved) common causes of the observed features (causal sufficiency).
3) The prediction is optimal, leveraging complete statistical information from the past of the features under consideration.
While practical exact satisfaction of these assumptions is challenging, Granger causality has found widespread application due to its simplicity. Our attribution analysis loosely aligns with Granger causality, as it hinges on comparing model performance with and without a specific feature.
Notably, our approach involves a post-hoc analysis of existing models, contrasting with the conventional method of training separate models with and without the feature for attributions.

As discussed in \cite{DBLP:conf/iclr/MakansiKLGJBS22}, caution must be exercised when drawing conclusions from small Shapley values, as their diminutiveness does not necessarily preclude the existence of causal influence; a model might not have learned to incorporate certain influences.
However, if a known causal influence exists, and the corresponding Shapley value approximates zero, it signals a deficiency in the model. Similarly, the model's detection of a causal influence aligns with the manifestation of non-zero Shapley values. In the specific task of stroke forecasting, where the influence of players is acknowledged on future strokes, albeit debatable for distant strokes, positive Shapley values still imply the presence of a causal influence.

%Bibliography
\bibliographystyle{icml2022}
\bibliography{main}

\begin{thebibliography}{37}
\providecommand{\natexlab}[1]{#1}
\providecommand{\url}[1]{\texttt{#1}}
\expandafter\ifx\csname urlstyle\endcsname\relax
  \providecommand{\doi}[1]{doi: #1}\else
  \providecommand{\doi}{doi: \begingroup \urlstyle{rm}\Url}\fi

\bibitem[Ancona et~al.(2019)Ancona, {\"{O}}ztireli, and Gross]{DBLP:conf/icml/AnconaOG19}
Ancona, M., {\"{O}}ztireli, C., and Gross, M.~H.
\newblock Explaining deep neural networks with a polynomial time algorithm for shapley value approximation.
\newblock In \emph{{ICML}}, volume~97 of \emph{Proceedings of Machine Learning Research}, pp.\  272--281. {PMLR}, 2019.

\bibitem[Bento et~al.(2021)Bento, Saleiro, Cruz, Figueiredo, and Bizarro]{DBLP:conf/kdd/0002SCFB21}
Bento, J., Saleiro, P., Cruz, A.~F., Figueiredo, M. A.~T., and Bizarro, P.
\newblock Timeshap: Explaining recurrent models through sequence perturbations.
\newblock In \emph{{KDD}}, pp.\  2565--2573. {ACM}, 2021.

\bibitem[Chang et~al.(2023)Chang, Wang, and Peng]{DBLP:journals/corr/abs-2211-12217}
Chang, K., Wang, W., and Peng, W.
\newblock Where will players move next? dynamic graphs and hierarchical fusion for movement forecasting in badminton.
\newblock In \emph{{AAAI}}, pp.\  6998--7005. {AAAI} Press, 2023.

\bibitem[Feng et~al.(2018)Feng, Wallace, Grissom~II, Iyyer, Rodriguez, and Boyd-Graber]{feng-etal-2018-pathologies}
Feng, S., Wallace, E., Grissom~II, A., Iyyer, M., Rodriguez, P., and Boyd-Graber, J.
\newblock Pathologies of neural models make interpretations difficult.
\newblock In \emph{Proceedings of the 2018 Conference on Empirical Methods in Natural Language Processing}, pp.\  3719--3728. Association for Computational Linguistics, October-November 2018.
\newblock \doi{10.18653/v1/D18-1407}.

\bibitem[Ge et~al.(2023)Ge, Song, and Huang]{DBLP:conf/aaai/GeSH23}
Ge, C., Song, S., and Huang, G.
\newblock Causal intervention for human trajectory prediction with cross attention mechanism.
\newblock In \emph{{AAAI}}, pp.\  658--666. {AAAI} Press, 2023.

\bibitem[Ghosh et~al.(2022)Ghosh, Ramamurthy, Chakma, and Roy]{DBLP:journals/percom/GhoshRCR22}
Ghosh, I., Ramamurthy, S.~R., Chakma, A., and Roy, N.
\newblock Decoach: Deep learning-based coaching for badminton player assessment.
\newblock \emph{Pervasive Mob. Comput.}, 83:\penalty0 101608, 2022.

\bibitem[Giuliari et~al.(2020)Giuliari, Hasan, Cristani, and Galasso]{DBLP:conf/icpr/GiuliariHCG20}
Giuliari, F., Hasan, I., Cristani, M., and Galasso, F.
\newblock Transformer networks for trajectory forecasting.
\newblock In \emph{25th International Conference on Pattern Recognition}, pp.\  10335--10342. {IEEE}, 2020.

\bibitem[Granger(1969)]{granger1969investigating}
Granger, C.~W.
\newblock Investigating causal relations by econometric models and cross-spectral methods.
\newblock \emph{Econometrica: journal of the Econometric Society}, pp.\  424--438, 1969.

\bibitem[Han et~al.(2020)Han, Wallace, and Tsvetkov]{DBLP:conf/acl/HanWT20}
Han, X., Wallace, B.~C., and Tsvetkov, Y.
\newblock Explaining black box predictions and unveiling data artifacts through influence functions.
\newblock In \emph{{ACL}}, pp.\  5553--5563. Association for Computational Linguistics, 2020.

\bibitem[Huang et~al.(2023)Huang, Hsueh, Chien, Wang, Wang, and Peng]{Huang_Hsueh_Chien_Wang_Wang_Peng_2023}
Huang, L., Hsueh, N., Chien, Y., Wang, W., Wang, K., and Peng, W.
\newblock A reinforcement learning badminton environment for simulating player tactics (student abstract).
\newblock In \emph{{AAAI}}, pp.\  16232--16233. {AAAI} Press, 2023.

\bibitem[Janzing et~al.(2013)Janzing, Balduzzi, Grosse-Wentrup, and Sch{\"o}lkopf]{janzing2013quantifying}
Janzing, D., Balduzzi, D., Grosse-Wentrup, M., and Sch{\"o}lkopf, B.
\newblock Quantifying causal influences.
\newblock 2013.

\bibitem[Janzing et~al.(2020)Janzing, Minorics, and Bl{\"{o}}baum]{DBLP:conf/aistats/JanzingMB20}
Janzing, D., Minorics, L., and Bl{\"{o}}baum, P.
\newblock Feature relevance quantification in explainable {AI:} {A} causal problem.
\newblock In \emph{{AISTATS}}, volume 108 of \emph{Proceedings of Machine Learning Research}, pp.\  2907--2916. {PMLR}, 2020.

\bibitem[Karami et~al.(2023)Karami, Lombaert, and Rivest{-}H{\'{e}}nault]{DBLP:journals/cmig/KaramiLR23}
Karami, M., Lombaert, H., and Rivest{-}H{\'{e}}nault, D.
\newblock Real-time simulation of viscoelastic tissue behavior with physics-guided deep learning.
\newblock \emph{Comput. Medical Imaging Graph.}, 104:\penalty0 102165, 2023.

\bibitem[Li et~al.(2016)Li, Monroe, and Jurafsky]{DBLP:journals/corr/LiMJ16a}
Li, J., Monroe, W., and Jurafsky, D.
\newblock Understanding neural networks through representation erasure.
\newblock \emph{CoRR}, abs/1612.08220, 2016.

\bibitem[Li et~al.(2023)Li, Liu, An, Qin, and Cheng]{DBLP:journals/sensors/LiLAQC23}
Li, W., Liu, X., An, K., Qin, C., and Cheng, Y.
\newblock Table tennis track detection based on temporal feature multiplexing network.
\newblock \emph{Sensors}, 23\penalty0 (3):\penalty0 1726, 2023.

\bibitem[Lien et~al.(2022)Lien, Lin, and Wang]{DBLP:conf/pkdd/LienLW22}
Lien, Y., Lin, Y., and Wang, Y.
\newblock Uncertainty awareness for predicting noisy stock price movements.
\newblock In \emph{{ECML/PKDD} {(6)}}, volume 13718 of \emph{Lecture Notes in Computer Science}, pp.\  154--169. Springer, 2022.

\bibitem[Liu \& Wang(2022)Liu and Wang]{DBLP:conf/cvpr/LiuW22}
Liu, P. and Wang, J.
\newblock Monotrack: Shuttle trajectory reconstruction from monocular badminton video.
\newblock In \emph{{CVPR} Workshops}, pp.\  3512--3521. {IEEE}, 2022.

\bibitem[Lundberg \& Lee(2017)Lundberg and Lee]{DBLP:conf/nips/LundbergL17}
Lundberg, S.~M. and Lee, S.
\newblock A unified approach to interpreting model predictions.
\newblock In \emph{{NIPS}}, pp.\  4765--4774, 2017.

\bibitem[Macquet \& Fleurance(2007)Macquet and Fleurance]{macquet2007naturalistic}
Macquet, A.-C. and Fleurance, P.
\newblock Naturalistic decision-making in expert badminton players.
\newblock \emph{Ergonomics}, 50\penalty0 (9):\penalty0 1433--1450, 2007.

\bibitem[Makansi et~al.(2022)Makansi, von K{\"{u}}gelgen, Locatello, Gehler, Janzing, Brox, and Sch{\"{o}}lkopf]{DBLP:conf/iclr/MakansiKLGJBS22}
Makansi, O., von K{\"{u}}gelgen, J., Locatello, F., Gehler, P.~V., Janzing, D., Brox, T., and Sch{\"{o}}lkopf, B.
\newblock You mostly walk alone: Analyzing feature attribution in trajectory prediction.
\newblock In \emph{{ICLR}}. OpenReview.net, 2022.

\bibitem[Niu et~al.(2021)Niu, Tang, Zhang, Lu, Hua, and Wen]{DBLP:conf/cvpr/NiuTZL0W21}
Niu, Y., Tang, K., Zhang, H., Lu, Z., Hua, X., and Wen, J.
\newblock Counterfactual {VQA:} {A} cause-effect look at language bias.
\newblock In \emph{{CVPR}}, pp.\  12700--12710. Computer Vision Foundation / {IEEE}, 2021.

\bibitem[Pearl(2009)]{pearl2009causality}
Pearl, J.
\newblock \emph{Causality}.
\newblock Cambridge university press, 2009.

\bibitem[Pearl et~al.(2016)Pearl, Glymour, and Jewell]{pearl2016causal}
Pearl, J., Glymour, M., and Jewell, N.~P.
\newblock \emph{Causal inference in statistics: A primer}.
\newblock John Wiley \& Sons, 2016.

\bibitem[Peters et~al.(2017)Peters, Janzing, and Sch{\"o}lkopf]{peters2017elements}
Peters, J., Janzing, D., and Sch{\"o}lkopf, B.
\newblock \emph{Elements of causal inference: foundations and learning algorithms}.
\newblock The MIT Press, 2017.

\bibitem[Shapley(1997)]{shapley1997value}
Shapley, L.~S.
\newblock A value for n-person games.
\newblock \emph{Classics in game theory}, 69, 1997.

\bibitem[Smilkov et~al.(2017)Smilkov, Thorat, Kim, Vi{\'{e}}gas, and Wattenberg]{DBLP:journals/corr/SmilkovTKVW17}
Smilkov, D., Thorat, N., Kim, B., Vi{\'{e}}gas, F.~B., and Wattenberg, M.
\newblock Smoothgrad: removing noise by adding noise.
\newblock \emph{CoRR}, abs/1706.03825, 2017.

\bibitem[Sun et~al.(2020{\natexlab{a}})Sun, Lin, Chuang, Hsu, Yu, Chung, and İk]{9302757}
Sun, N.-E., Lin, Y.-C., Chuang, S.-P., Hsu, T.-H., Yu, D.-R., Chung, H.-Y., and İk, T.-U.
\newblock Tracknetv2: Efficient shuttlecock tracking network.
\newblock In \emph{2020 International Conference on Pervasive Artificial Intelligence (ICPAI)}, pp.\  86--91, 2020{\natexlab{a}}.
\newblock \doi{10.1109/ICPAI51961.2020.00023}.

\bibitem[Sun et~al.(2020{\natexlab{b}})Sun, Davis, Schulte, and Liu]{DBLP:conf/kdd/SunDSL20}
Sun, X., Davis, J., Schulte, O., and Liu, G.
\newblock Cracking the black box: Distilling deep sports analytics.
\newblock In \emph{{KDD} '20: The 26th {ACM} {SIGKDD} Conference on Knowledge Discovery and Data Mining}, pp.\  3154--3162, 2020{\natexlab{b}}.

\bibitem[Sundararajan \& Najmi(2020)Sundararajan and Najmi]{DBLP:conf/icml/SundararajanN20}
Sundararajan, M. and Najmi, A.
\newblock The many shapley values for model explanation.
\newblock In \emph{{ICML}}, volume 119 of \emph{Proceedings of Machine Learning Research}, pp.\  9269--9278. {PMLR}, 2020.

\bibitem[Sundararajan et~al.(2017)Sundararajan, Taly, and Yan]{DBLP:conf/icml/SundararajanTY17}
Sundararajan, M., Taly, A., and Yan, Q.
\newblock Axiomatic attribution for deep networks.
\newblock In \emph{{ICML}}, volume~70 of \emph{Proceedings of Machine Learning Research}, pp.\  3319--3328. {PMLR}, 2017.

\bibitem[Vaswani et~al.(2017)Vaswani, Shazeer, Parmar, Uszkoreit, Jones, Gomez, Kaiser, and Polosukhin]{DBLP:conf/nips/VaswaniSPUJGKP17}
Vaswani, A., Shazeer, N., Parmar, N., Uszkoreit, J., Jones, L., Gomez, A.~N., Kaiser, L., and Polosukhin, I.
\newblock Attention is all you need.
\newblock In \emph{{NIPS}}, pp.\  5998--6008, 2017.

\bibitem[Wang(2022)]{DBLP:conf/cikm/Wang22}
Wang, W.
\newblock Modeling turn-based sequences for player tactic applications in badminton matches.
\newblock In \emph{{CIKM}}, pp.\  5128--5131. {ACM}, 2022.

\bibitem[Wang et~al.(2021)Wang, Chan, Yang, Wang, Fan, and Peng]{DBLP:conf/icdm/WangCYWFP21}
Wang, W., Chan, T., Yang, H., Wang, C., Fan, Y., and Peng, W.
\newblock Exploring the long short-term dependencies to infer shot influence in badminton matches.
\newblock In \emph{{ICDM}}, pp.\  1397--1402. {IEEE}, 2021.

\bibitem[Wang et~al.(2022)Wang, Shuai, Chang, and Peng]{DBLP:conf/aaai/WangSCP22}
Wang, W., Shuai, H., Chang, K., and Peng, W.
\newblock Shuttlenet: Position-aware fusion of rally progress and player styles for stroke forecasting in badminton.
\newblock In \emph{{AAAI}}, pp.\  4219--4227. {AAAI} Press, 2022.

\bibitem[Wang et~al.(2023{\natexlab{a}})Wang, Chan, Peng, Yang, Wang, and Fan]{10.1145/3551391}
Wang, W., Chan, T., Peng, W., Yang, H., Wang, C., and Fan, Y.
\newblock How is the stroke? inferring shot influence in badminton matches via long short-term dependencies.
\newblock \emph{{ACM} Trans. Intell. Syst. Technol.}, 14\penalty0 (1):\penalty0 7:1--7:22, 2023{\natexlab{a}}.

\bibitem[Wang et~al.(2023{\natexlab{b}})Wang, Du, and Peng]{ShuttleSet22}
Wang, W., Du, W., and Peng, W.
\newblock Shuttleset22: Benchmarking stroke forecasting with stroke-level badminton dataset.
\newblock \emph{CoRR}, abs/2306.15664, 2023{\natexlab{b}}.

\bibitem[Xu et~al.(2023)Xu, Tan, Tan, Chen, Wang, Wang, and Wang]{Xu_2023_CVPR}
Xu, C., Tan, R.~T., Tan, Y., Chen, S., Wang, Y.~G., Wang, X., and Wang, Y.
\newblock Eqmotion: Equivariant multi-agent motion prediction with invariant interaction reasoning.
\newblock In \emph{CVPR}, pp.\  1410--1420, 2023.

\end{thebibliography}

\end{document}